\documentclass[letterpaper, 10 pt, journal, twoside]{IEEEtran}

\IEEEoverridecommandlockouts                              % This command is only needed if 
\usepackage{graphics} % for pdf, bitmapped graphics files
\usepackage{epsfig} % for postscript graphics files
\usepackage{mathptmx} % assumes new font selection scheme installed
\usepackage{times} % assumes new font selection scheme installed
\usepackage{amsfonts} % blackboard math symbols
\usepackage{amssymb}  % assumes amsmath package installed
\usepackage{amsmath} % assumes amsmath package installed

\usepackage{graphicx}
\usepackage{helvet}

\usepackage{url}
\usepackage{rotating}
\usepackage{xspace}
\usepackage{booktabs}       % professional-quality tables
\usepackage{nccmath}
\usepackage{nicefrac}       % compact symbols for 1/2, etc.
\usepackage{microtype}      % microtypography
\usepackage{comment}		
\usepackage{adjustbox}		
\usepackage{tabu}
\usepackage{calc}			
\usepackage{wrapfig}

\usepackage{multirow}
\usepackage{float}
\usepackage[hang,flushmargin]{footmisc}
\usepackage[neverdecrease]{paralist}

\usepackage{pifont} %240228 추가 (체크표기)

\usepackage{makecell}
\usepackage{algorithm}
\usepackage{algorithmicx}
\usepackage{algpseudocode}

\usepackage{calc}
\usepackage{color}
\usepackage{colortbl}
\usepackage{subfigure}
\usepackage{animate}

\newcommand{\Secref}[1]{Section~\ref{#1}}
\newcommand{\figref}[1]{Fig.~\ref{#1}}
\newcommand{\tabref}[1]{Table~\ref{#1}}

\usepackage[mathcal]{eucal}
\usepackage{ntheorem}

\usepackage{tabularx} % preamble에 추가

\usepackage{hyperref}

\newcommand{\be}{\begin{eqnarray}}
\newcommand{\ee}{\end{eqnarray}}
\newcommand{\bee}{\begin{eqnarray*}}
\newcommand{\eee}{\end{eqnarray*}}

\newcommand{\matrixb}{\left[ \begin{array}}
\newcommand{\matrixe}{\end{array} \right]}

\definecolor{CuGray}{gray}{0.9}
\newcolumntype{g}{>{\columncolor{CuGray}}c}

\makeatletter
\DeclareRobustCommand\onedot{\futurelet\@let@token\@onedot}
\def\@onedot{\ifx\@let@token.\else.\null\fi\xspace}

\newcommand{\dashrule}[1][black]{%
  \color{#1}\rule[\dimexpr.5ex-.2pt]{4pt}{.4pt}\xleaders\hbox{\rule{4pt}{0pt}\rule[\dimexpr.5ex-.2pt]{4pt}{.4pt}}\hfill\kern0pt%
}

\title{\LARGE \bf
Flow4D: Leveraging 4D Voxel Network \\ for LiDAR Scene Flow Estimation
}
\author{Jaeyeul Kim$^{1}$, Jungwan Woo$^{1}$, Ukcheol Shin$^{2}$, Jean Oh$^{2}$, and Sunghoon Im$^{1}$% <-this % stops a space
\thanks{
(Corresponding author: Sunghoon Im.)}% <-this % stops a space
\thanks{\quad $^{1}$Jaeyeul Kim, Jungwan Woo, and Sunghoon Im are with the Department of Electrical Engineering and Computer Science, DGIST, Daegu, 42988, Republic of Korea (email: {\tt\small \{jykim94, friendship1, sunghoonim\}@dgist.ac.kr})}%
\thanks{\quad $^{2}$U. Shin and J. Oh are with Robotics Institute, Carnegie Mellon University, Pittsburgh, Pennsylvania, 15217, United States
        {\tt\small \{ushin, hyaejino\}@andrew.cmu.edu}}%
}

\begin{document}
%\markboth{IEEE Robotics and Automation Letters. Preprint Version. Accepted June, 2022}{Kim \MakeLowercase{\textit{et al.}}: RVMOS: Range-View Moving Object Segmentation}

\maketitle
\thispagestyle{empty}
\pagestyle{empty}

\begin{abstract}
Understanding the motion states of the surrounding environment is critical for safe autonomous driving.
These motion states can be accurately derived from scene flow, which captures the three-dimensional motion field of points.
Existing LiDAR scene flow methods extract spatial features from each point cloud and then fuse them channel-wise, resulting in the implicit extraction of spatio-temporal features.
Furthermore, they utilize 2D Bird’s Eye View and process only two frames, missing crucial spatial information along the Z-axis and the broader temporal context, leading to suboptimal performance.
To address these limitations, we propose Flow4D, which temporally fuses multiple point clouds after the 3D intra-voxel feature encoder, enabling more explicit extraction of spatio-temporal features through a 4D voxel network.
However, while using 4D convolution improves performance, it significantly increases the computational load.
For further efficiency, we introduce the Spatio-Temporal Decomposition Block (STDB), which combines 3D and 1D convolutions instead of using heavy 4D convolution.
In addition, Flow4D further improves performance by using five frames to take advantage of richer temporal information.
As a result, the proposed method achieves a 45.9\% higher performance compared to the state-of-the-art while running in real-time, and won 1\textsuperscript{st} place in the 2024 Argoverse 2 Scene Flow Challenge.
The code is available at \href{https://github.com/dgist-cvlab/Flow4D}{https://github.com/dgist-cvlab/Flow4D}.

\end{abstract}

%SceneFlow 중요성
% Autonomous vehicle (AV) 주변 3차원 환경을 정확히 인지하는 것은 안전한 자율주행을 위해 중요하며 많은 발전을 이루어왔다.
% 이러한 Perception task 중에서도, Traffic participants들의 Motion flow를 측정하는것은 AV의 경로 계획을 위해 필수적이다.
%그러나, 기존 SceneFlow
%이를 위해, 본 논문에서 우리는 LiDAR-based Scene Flow Framework인 Flow4D를 제안합니다.
%제안된 Flow4D는 다수의 시퀀셜 포인트클라우드들을 입력으로, 4D voxel space에서 인코더부터 직접적으로 spatio-temporal feature를 추출합니다.
%또한, Spatio-Temporal decomposition convolution을 제안하여 4D voxel 기반임에도 높은 연산효율을 보장합니다.
%이러한 방법을 통해 우리의 방법은 real-time에서 구동되면서도, state-of-the-art 방법을 00.0% 능가하며 2024 Argoverse Scene Flow Challenge에서 1위를 수상하였습니다. 
%소스코드는 공개할 예정입니다.

% \begin{IEEEkeywords}
% Autonomous driving, Moving Object Segmentation, Range-view, LiDAR, Perception
% \end{IEEEkeywords}

%%%%%%%%%%%%%%%%%%%%%%%%%%%%%%%%%%%%%%%%%%%%%%%%%%%%%%%%%%%%%%%%%%%%%%%%%%%%%%%%
% Introduction
\section{Introduction}
 \IEEEPARstart{L}iDAR provides accurate 3D distance measurements, making it widely used in autonomous vehicles (AVs). Given its precision and reliability, extensive research on LiDAR-based detection and semantic segmentation~\cite{li2023pillarnext, liu2024multi} has been conducted to perceive the static states of the environment.
These static states, such as the positions and sizes of objects and the semantic information of the surroundings, are crucial for understanding the driving environment.
However, for safe autonomous driving, information about the motion of surrounding objects is also essential.
Understanding these dynamic states allows AVs to avoid potential collisions and navigate more safely.
Due to this importance, LiDAR-based scene flow is emerging as a major topic in autonomous driving.

\begin{figure}[t] 
\centering
\footnotesize
\includegraphics[width=0.99\columnwidth]{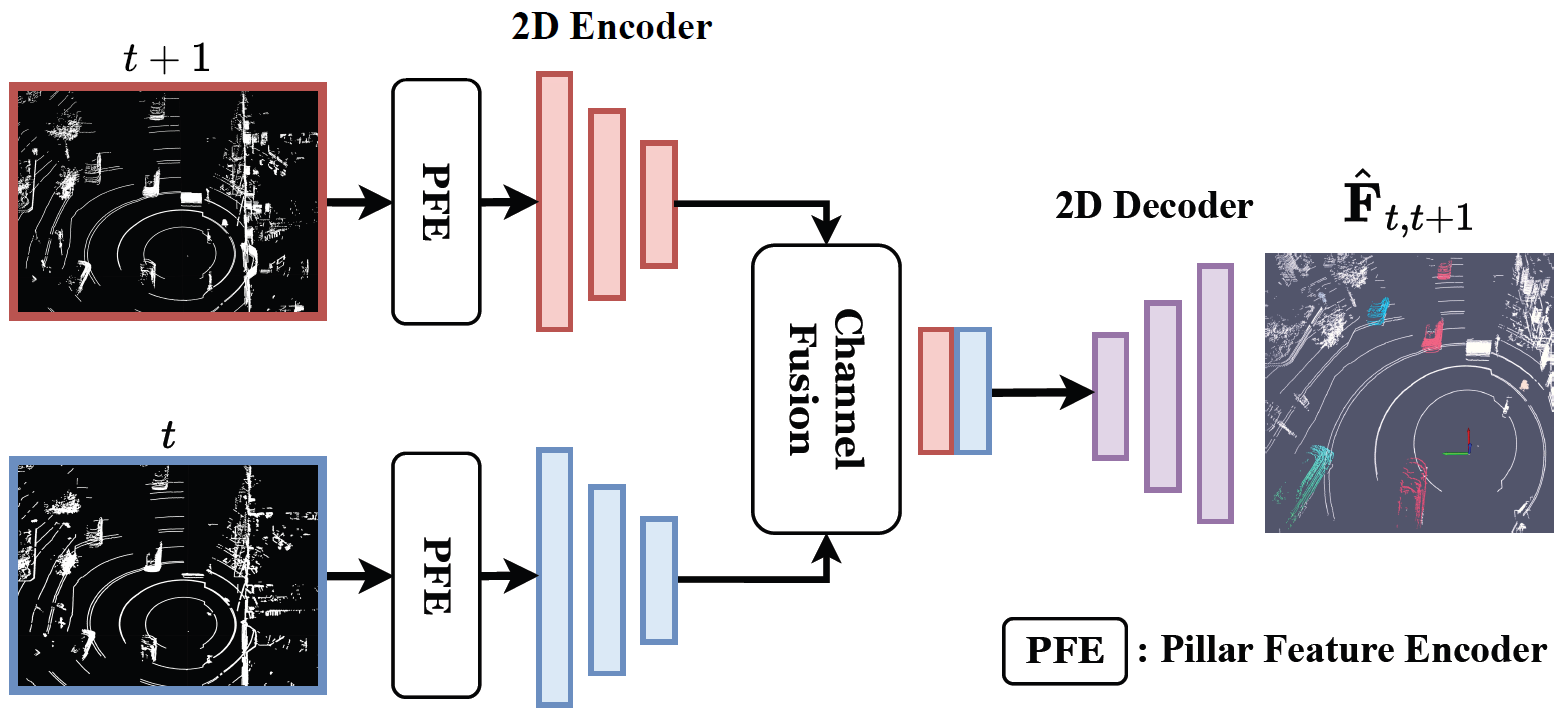}
%\par\vspace{-1mm}
\text{(a) Leveraging 2D Network~\cite{jund2022fastflow3d,zhang2024deflow}}
\par\vspace{2mm}
\includegraphics[width=0.99\columnwidth]{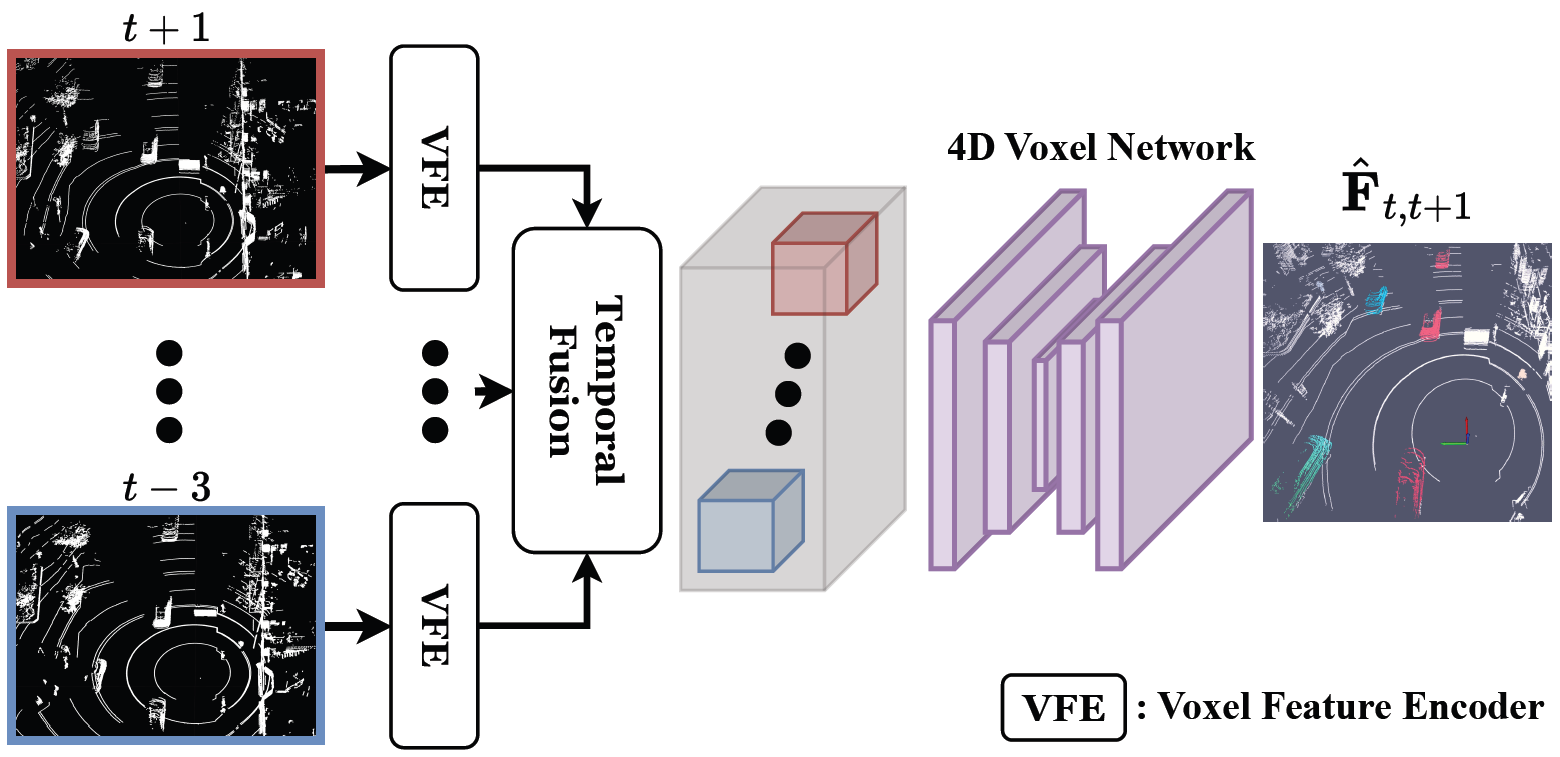}
%\par\vspace{-1mm}
\text{(b) Leveraging 4D Network (Ours)}
\par\vspace{2mm}
\caption{Fusion strategy comparison for LiDAR scene flow estimation.
% \jy{(a) Existing methods~\cite{jund2022fastflow3d,zhang2024deflow} use individual encoders for each point cloud and then fuse them channel-wise.
% (b) In contrast, our Flow4D merges five point clouds into 4D voxels temporally right after the VFE, enabling the explicit extraction of spatio-temporal features.}
(a) Existing methods~\cite{jund2022fastflow3d,zhang2024deflow} use a 2D pillar representation and find correspondences via channel fusion followed by a 2D decoder.
However, these methods only allow limited channel-level interaction due to their 2D representation.
(b) In contrast, our proposed method allows explicit spatio-temporal interaction to find accurate correspondences by using a 4D voxel representation.}
\label{fig_teaser}
\end{figure}

% (a) Traditional methods use individual encoders for each point cloud and then establish connections through the decoder. (b) Another approach uses separate detectors for each point cloud and estimates flow using a Kalman filter. Both methods extract spatial features separately before correlating them. (c) Our proposed Flow4D method employs an integrated 4D voxel-based network to extract spatio-temporal information directly from the encoder, significantly enhancing performance.
% (a) Previous methods~\cite{jund2022fastflow3d,zhang2024deflow} use individual encoders for each point cloud and establish connections through the decoder, leading to a late integration of features.
% In contrast, our Flow4D merges 3D point clouds into 4D voxels right after the voxel feature encoder, allowing spatio-temporal information to be extracted directly from the encoder, resulting in enhanced performance.

%기존 연구 얘기, + 그러나 한계
LiDAR-based scene flow studies aim to provide a detailed motion representation of the surrounding environment by estimating the 3D flow vector of each point.
Numerous self-supervised methods~\cite{li2021NSFP, li2023fastNSF, liu2024self, vedder2024zeroflow, lin2024icp} that do not require human-annotated data have been proposed, but they have not provided satisfactory performance.
In contrast, supervised methods~\cite{zhang2024deflow, khatri2024can} show improved results, thanks to the large-scale Argoverse 2~\cite{Benjamin2021Argoverse2} scene flow dataset.
However, most supervised methods also suffer from suboptimal performance because they fail to explicitly extract spatio-temporal features.
This is because they first extract spatial features from each point cloud and then fuse them channel-wise to obtain temporal correlations, as shown in \figref{fig_teaser}-(a).
In addition, they use only two frames, missing the richer temporal context available in additional past frames.

To overcome these limitations, we propose Flow4D, a straightforward and highly effective LiDAR-based scene flow framework that temporally fuses point clouds and uses a 4D voxel network to explicitly extract spatio-temporal features. 
We first 3D voxelize each point cloud with a simple intra-voxel feature encoder (VFE).
After that, by adding a time dimension to each 3D voxel and then concatenating along this new axis to form a 4D voxel space, the spatio-temporal features can be directly extracted.
While 4D convolution can explicitly extract spatial and temporal features simultaneously, it is computationally intensive.
To solve this issue, we propose the Spatio-Temporal Decomposition Block (STDB), which separates 4D convolution into 3D spatial and 1D temporal convolutions, enabling more efficient computation.
Additionally, owing to its early fusion design that fuses point clouds immediately after the VFE, our method can efficiently utilize more point clouds, leveraging their richer temporal information.
As a result, our Flow4D achieves superior performance compared to existing state-of-the-art methods while maintaining high computational efficiency that allows for real-time operation.

% Next, each 3D voxel is unsqueezed to add an additional time dimension, and then they are concatenated along this new dimension to form a 4D voxel space, from which spatio-temporal features can be directly extracted.

% Additionally, the early fusion design, which merges point clouds immediately after the VFE, enables the efficient utilization of more point clouds, thereby leveraging richer temporal cues.

%240616
%To overcome these issues, we propose Flow4D, a simple yet highly effective early-fusion strategy.
% Moreover, unlike the late-fusion methods that require heavy encoders or networks for each input point cloud, our method minimizes the computational load from the increased number of input point clouds.

% This approach does not significantly increase the computational load as the number of input point clouds increases, allowing our method to use multiple point clouds as input to extract richer features.
%% To address these limitations, we propose 4D voxel-based Flow4D, which effectively exploits spatio-temporal features from sequential point clouds.

%Summary
Our contributions can be summarized as follows:
\begin{itemize}
    \item We present Flow4D, a simple yet highly effective 4D voxel-based scene flow framework that explicitly extracts spatio-temporal information from multiple point clouds.
    \item We introduce the Spatio-Temporal Decomposition Block (STDB), which enhances computational efficiency by decomposing 4D convolution into 3D spatial and 1D temporal convolutions.
    % \item We design Flow4D with an early fusion strategy, optimizing the utilization of multiple point clouds and thereby leveraging richer temporal cues.
    % \item The proposed method outperforms the existing state-of-the-art by 45.9\% on the Argoverse 2 dataset, while operating in real-time on a desktop GPU.
    \item The proposed method outperforms the existing state-of-the-art by 45.9\% on the Argoverse 2 dataset, running at a real-time speed of 15.1 FPS on an RTX 3090 GPU.
 
\end{itemize}

    % \item We present Flow4D, a novel scene flow framework that effectively extracts spatio-temporal information from 4D voxel space using multiple sequential point clouds.

% Related Works
\section{Related Work}

\subsection{LiDAR Data Processing} 
LiDAR point clouds are unordered, unstructured, and irregular, making the direct application of convolution operations difficult.
Therefore, to process point clouds, three methods are typically used: projection-based, point-based, and voxel-based approaches.
Firstly, projection-based methods convert 3D point clouds into a 2D grid form, such as Bird's Eye View (BEV)~\cite{Lang2019Pointpillars, li2023pillarnext} or Range View (RV)~\cite{Cortinhal2020SalsaNext, Ando2023Rangevit}. These methods have the advantage of efficiently compressing 3D data into 2D data. However, BEV may lose detailed information on the z-axis and RV suffer from distortion of receptive field.
Secondly, point-based methods~\cite{Thomas2019KPConv, Hu2020RandLA} directly extract features from raw point clouds, thus avoiding data distortion issues.
However, their computational efficiency can be a bit lower when applied to large-scale 3D data.
Lastly, voxel-based methods~\cite{Choyi2019MINKOWSKI, Zhu2021Cylindrical} convert point clouds into a regular grid of voxels, enabling the use of efficient 3D or 4D convolution operations.
Combined with sparse convolutions~\cite{liu2015sparse, graham2017submanifold}, voxel-based methods can efficiently operate while preserving geometric information.
% Although 3D convolutions are computationally intensive, applying 3D sparse convolutions~\cite{liu2015sparse, graham2017submanifold} allows for efficient processing of point clouds.

\subsection{LiDAR-based Self-supervised Scene Flow Estimation} 
Labeling point-wise scene flow in 3D LiDAR point clouds is very costly, which has led to the development of various self-supervised studies~\cite{mittal2020just, pontes2020scene, kittenplon2021flowstep3d, li2021NSFP, li2022rigidflow, li2023fastNSF, liu2024self, vedder2024zeroflow, lin2024icp, zhang2024seflow}.
Li et al.~\cite{li2021NSFP} propose a method that uses a neural scene flow prior to regularize the scene flow problem by leveraging runtime optimization, enabling better generalization to new environments.
FastNSF~\cite{li2023fastNSF} replaces the computationally expensive Chamfer loss with a distance transform-based loss in a runtime optimization-based neural scene flow pipeline.
Liu et al.~\cite{liu2024self} present a self-supervised multi-frame scene flow estimation method that leverages temporal information from previous point cloud frames to enhance accuracy and generalization performance.
ZeroFlow~\cite{vedder2024zeroflow} leverages label-free optimization~\cite{li2021NSFP} to generate pseudo-labels, supervising feedforward model~\cite{jund2022fastflow3d} without human annotations. This approach combines the efficiency of feedforward method with the robustness of optimization technique, enabling scalable and fast scene flow estimation.
Lin et al.~\cite{lin2024icp} propose ICP-Flow, a learning-free scene flow estimator that leverages the Iterative Closest Point (ICP) algorithm with a histogram-based initialization.
SeFlow~\cite{zhang2024seflow} employs DUFOMap to classify point clouds into dynamic and static points and utilizes HDBSCAN to cluster the dynamic points.
However, as large-scale human-annotated datasets have emerged, these self-supervised studies face the limitation of inferior performance compared to the latest supervised methods.

% \begin{figure}[t] 
% \centering
% \includegraphics[width=0.6\columnwidth]{figure_icra25/dataset_temp2.png}
% \caption{Argoverse 2 scene flow dataset.} 
% \label{fig_dataset}
% \end{figure}

\subsection{LiDAR-based Supervised Scene Flow Estimation} 
With the advent of the human-annotated scene flow datasets, various supervised scene flow methods~\cite{behl2019pointflownet, liu2019flownet3d, puy2020flot, wu2020pointpwc, lee2020pillarflow, li2021hcrf, jund2022fastflow3d, battrawy2022rms, zhang2024deflow, khatri2024can} have emerged.
Jund et al.~\cite{jund2022fastflow3d} introduce FastFlow3D, an efficient approach that converts 3D point clouds into a 2D Bird's Eye View (BEV) using pillar encoding~\cite{Lang2019Pointpillars}.
They employ a U-Net style 2D network to enable real-time flow estimation even with large-scale LiDAR data.
DeFlow~\cite{zhang2024deflow} addresses the loss of point-wise features during the voxelization process by refining voxel features and point features through a Gated Recurrent Unit.
Additionally, they propose a novel loss function that dynamically adjusts weights based on the motion states of each point.
Khatri et al.~\cite{khatri2024can} point out that traditional metrics fail to properly reflect the performance on small objects.
They propose a new class-aware and speed-normalized metric called Bucket Normalized Endpoint Error (EPE).
Additionally, they introduce TrackFlow, which uses a class-aware off-the-shelf detector and a 3D Kalman filter to estimate 3D scene flow.
While previous methods~\cite{jund2022fastflow3d, zhang2024deflow, khatri2024can} use only two point clouds, extracting spatial features from each and obtaining correlations through a decoder or Kalman filter, our Flow4D simultaneously extracts spatial and temporal features from multiple point clouds using a 4D voxel network.
% Previous methods~\cite{jund2022fastflow3d, zhang2024deflow, khatri2024can} use only two point clouds, extracting spatial features from each point cloud and obtaining correlations through a decoder or Kalman filter.
% In contrast, our Flow4D uses a 4D voxel-based network to simultaneously extract spatial and temporal features from multiple point clouds.

% Methodology

\begin{figure*}[t] 
	\centering
    \includegraphics[width=0.99\linewidth]{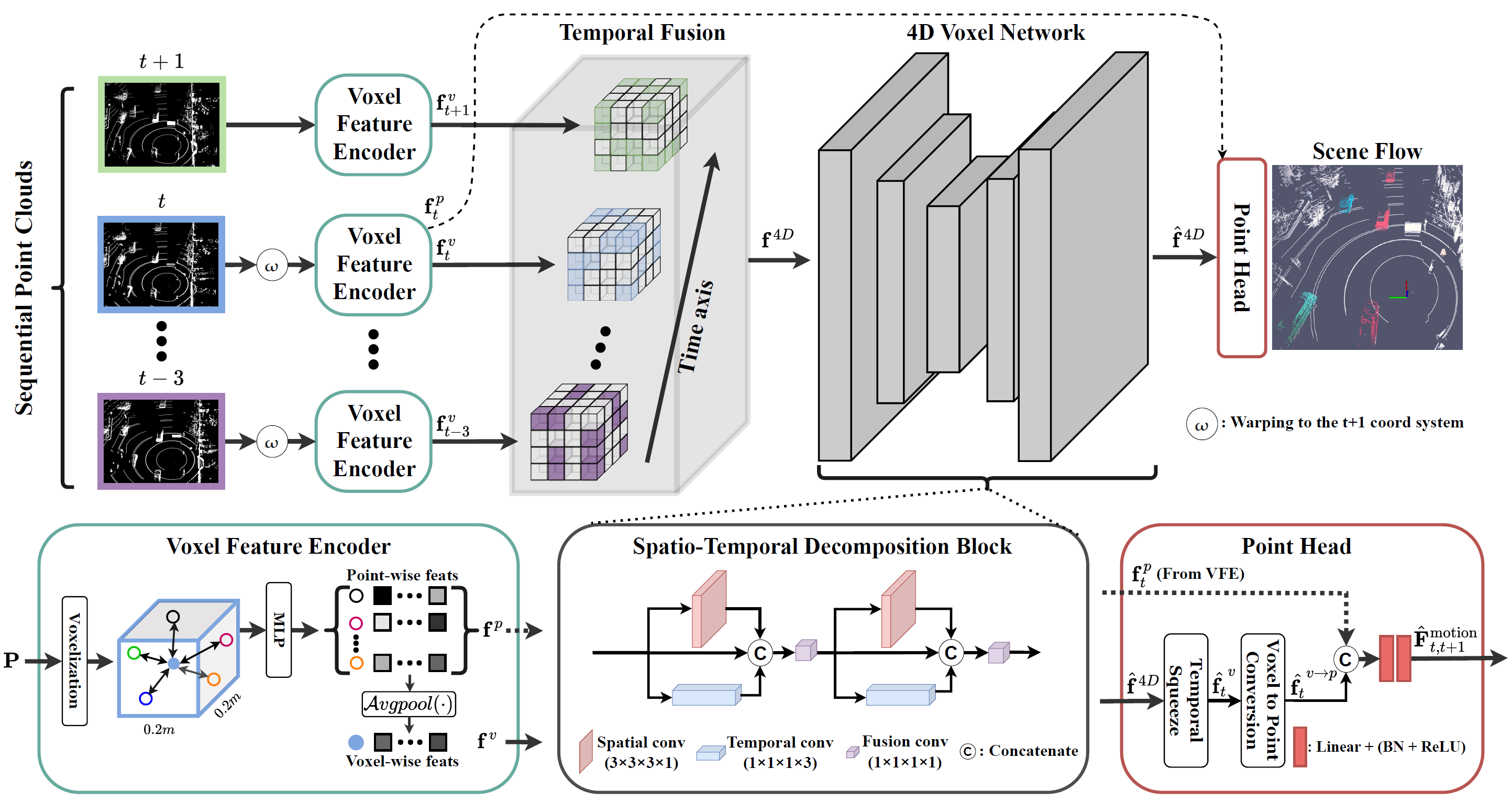}
    % \includegraphics[width=0.99\linewidth]{figure/main_fig_temp_jy.PNG}
	%\caption{Overall framework of the Flow4D.}
 \caption{Overall framework of the Flow4D. Sequential point clouds are processed through individual Voxel Feature Encoders to extract intra-voxel features. These features are then temporally fused to form 4D voxel features. The 4D voxel network, which consists of Spatio-Temporal Decomposition Blocks (STDBs), extracts voxel-wise spatio-temporal features. Finally, the Point Head module estimates point-wise flow vectors.}
	\label{fig_overview}
\end{figure*}

\section{Proposed Method}
In this section, we present Flow4D, a 4D voxel-based scene flow framework, with an overview provided in~\figref{fig_overview}.
First, in~\Secref{sec:overview}, we provide a brief overview of the 3D scene flow task.
In~\Secref{sec:4D_voxelization}, we introduce a method for creating 4D voxel representations by temporally fusing sequential point clouds.
\Secref{sec:4D_Network} and~\Secref{sec:head} explain the 4D voxel network and the point head, respectively.
% Finally, we describe the training details of the proposed method in~\Secref{sec:details}.

\subsection{LiDAR-based Scene Flow} %Problem definition, Overview
\label{sec:overview}
3D scene flow task aims to estimate the flow vector $\mathbf{F}_{t, t+1} \in \mathbb{R}^{N_{t} \times 3}$ from the current point cloud $\mathbf{P}_t \in \mathbb{R}^{N_{t} \times 3}$ to the next time step point cloud $\mathbf{P}_{t+1} \in \mathbb{R}^{N_{t+1} \times 3}$, where $N_t$ represents the number of points at time step $t$.
% Scene flow $\mathbf{F}_{t, t+1}$ can be decomposed into the ego-vehicle's motion $\mathbf{F}_{t, t+1}^{\text{ego}}$ and the motion vector of individual points $\mathbf{F}_{t, t+1}^{\text{motion}}$.
The scene flow vector $\mathbf{F}_{t, t+1}$ can be decomposed into the ego-vehicle's motion and the motion vector of individual points, as follows:
\begin{equation}
\mathbf{F}_{t, t+1} = \mathbf{F}_{t, t+1}^{\text{ego}} + \mathbf{F}_{t, t+1}^{\text{motion}}.
\end{equation}
Following previous studies~\cite{jund2022fastflow3d, zhang2024deflow, vedder2024zeroflow, khatri2024can}, we assume that $\mathbf{F}_{t, t+1}^{\text{ego}}$ is given and aim to obtain $\mathbf{F}_{t, t+1}^{\text{motion}}$.

%3D Scene flow는 현재 시점 포인트 클라우드 P_t에서 P_t+1로 향하는 flow vector F_t,t+1을 추정하는 것을 목표로 합니다.
%F는 ego-vehicle의 움직임으로 인한 F_ego와 각 포인트들의 motion vector인 F_motion으로 나타낼 수 있습니다.
%우리는 기존 연구들에 따라 F_ego는 주어졌다고 가정하고, F_motion을 획득하는 것을 목표로 합니다.

\subsection{Temporal Fusion} %=4D voxelization
\label{sec:4D_voxelization}
Existing studies predict motion vectors \(\hat{\mathbf{F}}_{t, t+1}^{\text{motion}}\) using only \(\mathbf{P}_t\) and \(\mathbf{P}_{t+1}\), but we find that leveraging additional past point clouds provides richer temporal cues to improve the accuracy of current flow prediction.
In this paper, we use five consecutive point clouds \(\mathbf{P}_{\tau}\) from the time steps \( \tau \):
\begin{equation}
\tau \in \{t-3, t-2, t-1, t, t+1\}.
\end{equation}
First, following previous studies~\cite{jund2022fastflow3d, zhang2024deflow}, we warp all point clouds to the coordinate system at \( t+1 \) time step using known poses as follows:
\begin{equation}
\mathbf{P}_{\tau \rightarrow t+1} = \mathbf{T}_{\tau, t+1} \mathbf{P}_{\tau}, \quad \forall \tau \ne t+1,
\end{equation}
where \(\mathbf{T}_{\tau, t+1}\) is the transformation matrix from time step \( \tau \) to \( t+1 \).

% Existing studies predict \(\hat{\mathbf{F}}_{t, t+1}\) using only \(\mathbf{P}_t\) and \(\mathbf{P}_{t+1}\), but we find that the additional use of past point clouds can be helpful in predicting current flow.

% The inclusion of additional past point clouds provides richer temporal context, allowing for more accurate and robust flow predictions.

% \begin{equation}
% t_i \in \{t-3, t-2, t-1, t, t+1\}.
% \end{equation}

% First, following previous studies~\cite{jund2022fastflow3d, zhang2024deflow}, we warp all point clouds to the coordinate system at \( t+1 \) time step using known poses as follows:
% \begin{equation}
% \mathbf{P}_{t_i \rightarrow t+1} = \mathbf{T}_{t_i, t+1} \mathbf{P}_{t_i}, \quad \forall t_i \ne t+1,
% \end{equation}
% where \(\mathbf{T}_{t_i, t+1}\) is the transformation matrix from time step \( t_i \) to \( t+1 \).

%%%%%%
%
%여기에 추가작성 2D -> 3D

Previous methods then convert these continuous point clouds into discrete 2D Bird’s Eye View representations.
However, this process compresses the Z-axis dimension, leading to a loss of crucial spatial information. 
To overcome this limitation, we employ 3D voxelization to retain critical spatial information along the Z-axis.
%0705t_i -> tau
First, we divide the 3D point cloud into a 3D grid and then extract intra-voxel features using a lightweight Voxel Feature Encoder (VFE).
Following~\cite{Lang2019Pointpillars}, for each point cloud $\mathbf{P}_{\tau} \in \mathbb{R}^{N_{\tau} \times 3}$, we concatenate the 3D coordinates of each point cloud $(x,y,z)$, the offsets from voxel center $(x-x_v, y-y_v, z-z_v)$, and the offsets from cluster center $(x-x_c, y-y_c, z-z_c)$ to form a tensor of shape $\mathbb{R}^{N_{\tau} \times 9}$.
By applying linear layers followed by Batch Normalization and ReLU, we obtain the initial point-wise features  \(\mathbf{f}_{\tau}^{p} \in \mathbb{R}^{N_{\tau} \times 16}\).
The initial voxel features $\mathbf{f}_{\tau}^{v} \in \mathbb{R}^{W \times L \times H \times 16}$ are obtained by performing voxel-wise average pooling on the $\mathbf{f}_{\tau}^{p}$.
We expand $\mathbf{f}_{\tau}^{v}$ by adding a time step dimension, resulting in a tensor of shape $\mathbb{R}^{W \times L \times H \times 1 \times 16}$.
Then, by concatenating these tensors, which are derived from five consecutive frames, along the time step axis, we obtain $\mathbf{f}^{4D} \in \mathbb{R}^{W \times L \times H \times 5 \times 16}$.
Thanks to this early fusion strategy, where fusion is performed right after the VFE for each point cloud, our method can effectively utilize a larger number of frames.

\subsection{4D Voxel Scene Flow Network}
\label{sec:4D_Network}
To directly extract spatio-temporal features, we design a 4D voxel network with an hourglass architecture incorporating skip connections.
% Since 2D convolution kernels primarily focus on spatial relationships within the height and width dimensions, temporal features are implicitly learned.
% In contrast, 3D or 4D convolutions, which include the time dimension, directly process both spatial and temporal dimensions at the kernel level, allowing explicit extraction of spatio-temporal information.
% While 2D convolution kernels primarily focus on spatial relationships within the height and width dimensions, implicitly learning temporal features, 3D or 4D convolutions include the time dimension and directly process both spatial and temporal dimensions at the kernel level, allowing for explicit extraction of spatio-temporal information.
2D convolution kernels primarily focus on spatial relationships within the height and width dimensions, implicitly extracting temporal features.
In contrast, 3D and 4D convolutions, including the time dimension, directly process both spatial and temporal information at the kernel level.
This allows for explicit extraction of spatio-temporal features.
Although using 4D convolutions ensures high performance, it also incurs significant computational overhead.
To address this issue, we propose the Spatio-Temporal Decomposition Block (STDB), which achieves similar performance to 4D convolution while operating more efficiently.
%with less memory and fewer FLOPs.
We replace the 4D convolution layer with a kernel size of $3 \times 3 \times 3 \times 3$ with a combination of a spatial convolution layer of $3 \times 3 \times 3 \times 1$ and a temporal convolution layer of $1 \times 1 \times 1 \times 3$.
These layers operate on the 4D voxel as 3D and 1D convolutions, respectively, allowing for reduced memory usage and increased computational speed compared to the baseline using 4D convolution.
% As shown in~\figref{fig_decomposition},
We introduce two types of decomposition blocks to replace the 4D convolution residual block.

%0617
% To enable the explicit extraction of both spatial and temporal features simultaneously, we design 4D voxel-based scene flow network.
% Using 3D or 4D convolution with a time dimension allows for the explicit extraction of spatio-temporal information because it processes spatial dimensions and the temporal dimension simultaneously.
% In contrast, 2D convolution kernels primarily focus on spatial relationships within the height and width dimensions, meaning that temporal cues are implicitly learned. 
% However, while the use of 4D convolution ensures high performance, it comes with significant computational overhead.
% We use \(\mathbf{f}^{4D}\) as the input to the 4D voxel-based network, which outputs the encoded 4D voxel-wise features \(\hat{\mathbf{f}}^{4D} \in \mathbb{R}^{W \times L \times H \times 5 \times 16}\).
% \noindent\textbf{(a) Sequential Spatio-temporal Block:} First, we design a simple method of sequentially decomposing each 4D conv layer into a combination of a 3D spatial conv layer and a 1D temporal conv layer.
% \noindent\textbf{(a) Sequential Decomposition Block:} First, we simply decompose each 4D conv layer into a combination of a 3D spatial conv layer and a 1D temporal conv layer.
\noindent\textbf{Parallel Spatio-Temporal Decomposition Block:}
First, we propose a parallel decomposition approach, where a 3D spatial convolution and a 1D temporal convolution are applied in parallel, as shown in~\figref{fig_overview}.
The 3D spatial convolution captures spatial features, while the 1D temporal convolution processes the time step dimension, focusing on temporal cues.
Their outputs are fused using a 1D convolution, combining spatial and temporal features into a unified representation.
This parallel approach efficiently extracts and integrates spatial and temporal features, reducing computational overhead.

\begin{figure}[t]
     \centering
     \footnotesize
     \includegraphics[width=0.8\columnwidth]{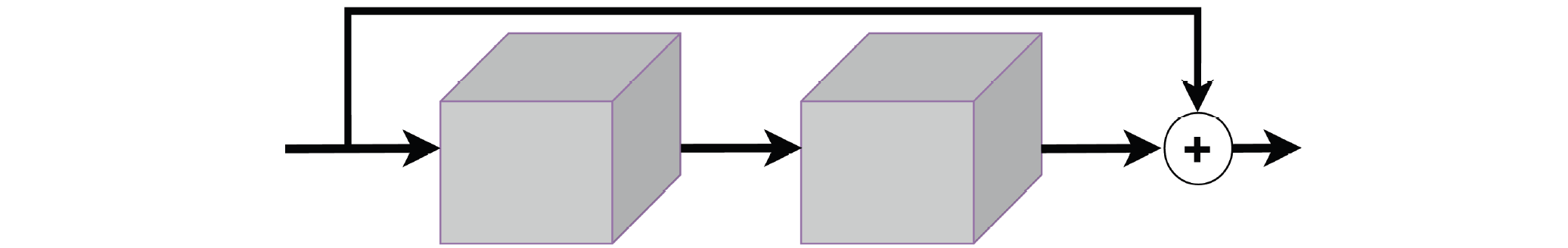}
    %\par\vspace{1mm}
    \text{(a) 4D convolution residual block}
    \par\vspace{3mm}
     \includegraphics[width=0.8\columnwidth]{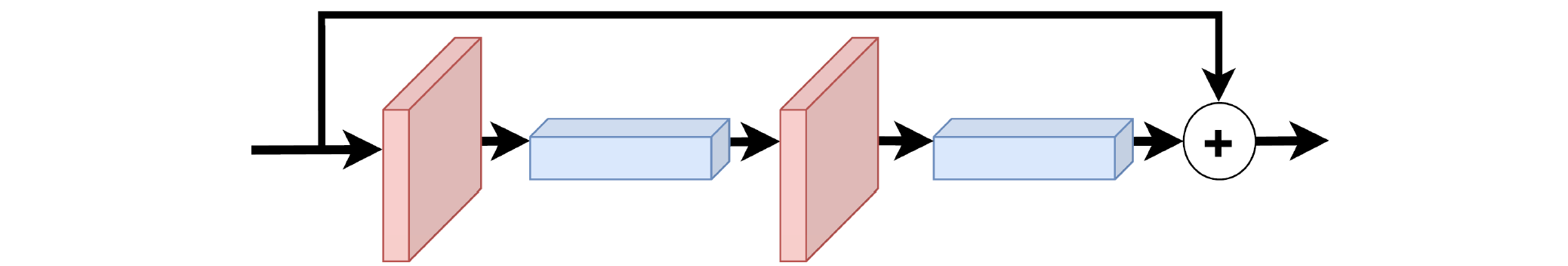}
    \par\vspace{0.0mm}
    \text{(b) Baseline spatio-temporal decomposition block (STDB-B)}
    \par\vspace{2.5mm}
     \includegraphics[width=0.8\columnwidth]{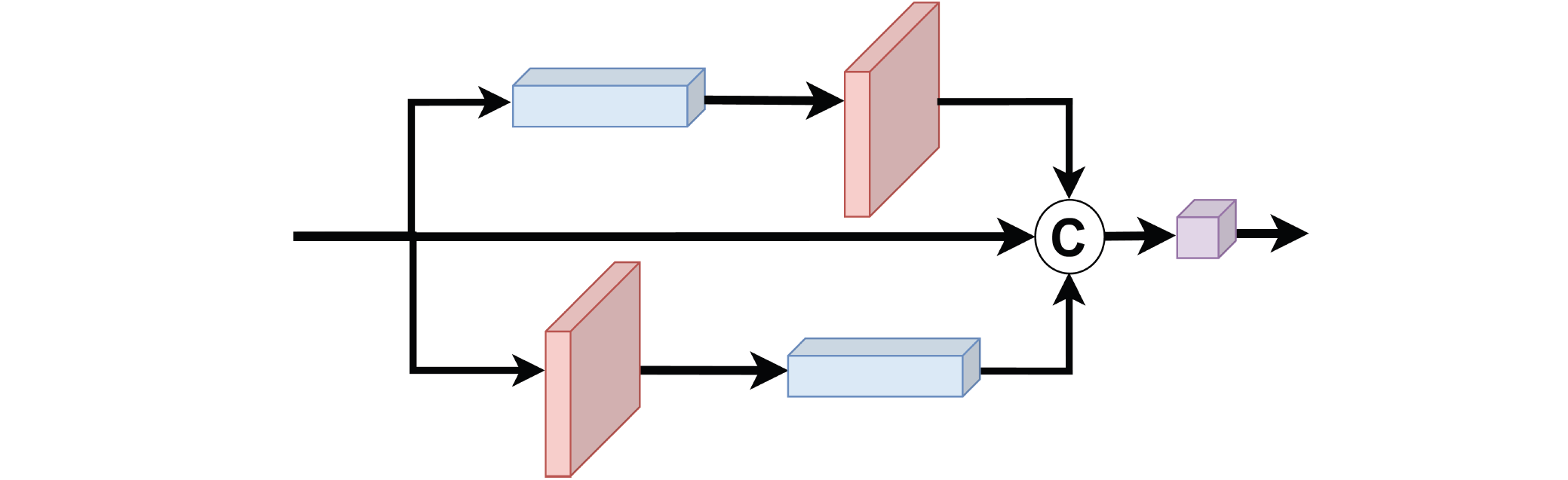}
    \par\vspace{-0.5mm}
    \text{(c) Dual-path spatio-temporal decomposition block (STDB-D)}
    \par\vspace{2mm}
     \includegraphics[width=0.8\columnwidth]{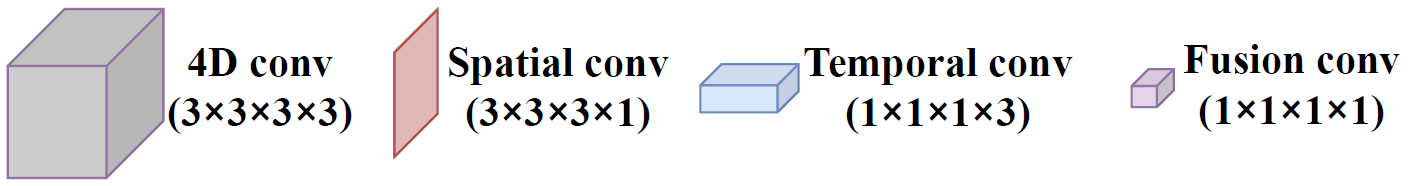}
     \caption{Comparison of spatio-temporal feature extraction blocks.}
     \label{fig_decomposition}
 \end{figure}

\noindent\textbf{Dual-path Spatio-Temporal Decomposition Block:} Secondly, we present a dual-path decomposition approach, as seen in~\figref{fig_decomposition}-(c).
In the top branch, a 3D spatial convolution is followed by a 1D temporal convolution.
Conversely, in the bottom branch, a 1D temporal convolution is followed by a 3D spatial convolution.
The outputs from these two branches are then fused using a 1D convolution.
This dual-path decomposition block extracts spatial and temporal features in different sequences before combining them, enhancing the efficiency and flexibility of feature extraction.

%0617
% First, we propose a parallel decomposition approach, where a 3D spatial convolution and a 1D temporal convolution are applied in parallel.
% Their outputs are then fused using a 1D convolution.
% This parallel approach ensures efficient extraction and integration of spatial and temporal features.
We design the 4D voxel network by applying the Spatio-Temporal Decomposition Block (STDB) twice at each level of the encoder and once at each level of the decoder.
We utilize submanifold sparse convolution~\cite{graham2017submanifold} for all 1D, 3D, and 4D convolutions to enhance computational efficiency.
The details of the network are provided in \tabref{tab:network}.

% \begin{table}[]
% \caption{\textbf{Network details.}}
% \centering
% \resizebox{0.99\columnwidth}{!}{
% \begin{tabular}{lcccc}
% \hline
%                          & Lv. & STCD & Pool \& Upsample     & Output shape \\ \hline
% Input                    &          &      &           & 512x512x32x5x16            \\ \hline
% \multirow{5}{*}{Encoder} & 1        & x2   & (2,2,2,1) & 256x256x16x5x32            \\ \cline{2-5} 
%                          & 2        & x2   & (2,2,2,1) & 128x128x8x5x64             \\ \cline{2-5} 
%                          & 3        & x2   & (2,2,2,1) & 64x64x4x5x64               \\ \cline{2-5} 
%                          & 4        & x2   & (2,2,1,1) & 32x32x4x5x64               \\ \cline{2-5} 
%                          & 5        & x2   & -         & 32x32x4x5x64               \\ \hline
% \multirow{5}{*}{Decoder} & 6        & x1   & (2,2,1,1) & 64x64x4x5x64               \\ \cline{2-5} 
%                          & 7        & x1   & (2,2,2,1) & 128x128x8x5x64             \\ \cline{2-5} 
%                          & 8        & x1   & (2,2,2,1) & 256x256x16x5x64            \\ \cline{2-5} 
%                          & 9        & x1   & (2,2,2,1) & 512x512x32x5x32            \\ \cline{2-5} 
%                          & 10       & x1   & -         & 512x512x32x5x16            \\ \hline
% \end{tabular}
% }
% \end{table}

\begin{table}[]
\caption{\textbf{Details of 4D voxel network.} Each STDB-P consists of two sets of convolution layers (spatial, temporal, and fusion conv). The values (x, y) in the STDB-P column represent the number of filters for each of these sets. Each stage performs pooling or upsampling after the STDB-P.}
\centering
\resizebox{0.99\columnwidth}{!}{
\begin{tabular}{ccccc}
\hline
                         & Stage & \makecell{STDB-P \\ (Filters)}                                & \makecell{Pool \& Up \\ (Kernel, Stride)} & \makecell{Output shape \\  \texttt{[W$\times$L$\times$H$\times$T$\times$ch]}} \\ \hline
Input                    &     &                                     &                                           & 512$\times$512$\times$32$\times$5$\times$16 \\ \hline
\multirow{9}{*}{Encoder} & 1   & \makecell{(16, 32) \\ (32, 32)} & Pool (2,2,2,1)                                & 256$\times$256$\times$16$\times$5$\times$32 \\ \cline{2-5} 
                         & 2   & \makecell{(32, 64) \\ (64, 64)} & Pool (2,2,2,1)                                & 128$\times$128$\times$8$\times$5$\times$64  \\ \cline{2-5} 
                         & 3   & \makecell{(64, 64) \\ (64, 64)} & Pool (2,2,2,1)                                & 64$\times$64$\times$4$\times$5$\times$64    \\ \cline{2-5} 
                         & 4   & \makecell{(64, 64) \\ (64, 64)} & Pool (2,2,1,1)                                & 32$\times$32$\times$4$\times$5$\times$64    \\ \cline{2-5} 
                         & 5   & \makecell{(64, 64) \\ (64, 64)} & Up (2,2,1,1)                                         & 32$\times$32$\times$4$\times$5$\times$64    \\ \hline
\multirow{4}{*}{Decoder} & 6   & (64, 64)                      & Up (2,2,2,1)                                & 64$\times$64$\times$4$\times$5$\times$64    \\ \cline{2-5} 
                         & 7   & (64, 64)                      & Up (2,2,2,1)                                & 128$\times$128$\times$8$\times$5$\times$64  \\ \cline{2-5} 
                         & 8   & (64, 64)                      & Up (2,2,2,1)                                & 256$\times$256$\times$16$\times$5$\times$64 \\ \cline{2-5} 
                         & 9   & (32, 16)                      & -                                & 512$\times$512$\times$32$\times$5$\times$16 \\  \hline
\end{tabular}
}
\label{tab:network}
\end{table}

\subsection{Point Head}
\label{sec:head}
Since the 4D voxel network outputs only inter-voxel features, we use a simple point head to output point-wise scene flow vector at the current time step $t$.
% First, we convert the 4D voxel-wise features $\hat{\mathbf{f}}^{4D}$ extracted from the 4D voxel-based network into point-wise features at the current time step $t$.
%\in \mathbb{R}^{W \times L \times H \times 5 \times 16}$.
First, we extract the features at the current time step $t$ from the output 5D tensor $\hat{\mathbf{f}}^{4D} \in \mathbb{R}^{W \times L \times H \times 5 \times 16}$, which is the output of the 4D voxel network, and then squeeze the time step dimension to form a 4D tensor $\hat{\mathbf{f}_t}^v \in \mathbb{R}^{W \times L \times H \times 16}$.
Then, we convert these voxel-wise features $\hat{\mathbf{f}_t}^v$ into point-wise features $\hat{\mathbf{f}_t}^{v \rightarrow p} \in \mathbb{R}^{N_t \times 16}$.
We concatenate $\hat{\mathbf{f}_t}^{v \rightarrow p}$ with the initial point-wise features $\mathbf{f}_t^{p}$ generated through voxel feature encoder.
Finally, we utilize two MLP layers to output the point-wise motion vectors $\mathbf{\hat{F}}_{t, t+1}^{\text{motion}} \in \mathbb{R}^{N_{t} \times 3}$.

\begin{table*}[t]
\caption{\textbf{Quantitative comparison on the Argoverse 2 test set.} `Sup.' indicates supervised methods and `\# F.' represents the number of input frames. `FD' represents Foreground Dynamic, `BS' represents Background Static, and `FS' represents Foreground Static.}
\centering
\resizebox{0.99\textwidth}{!}{
\begin{tabular}{l|c|c|ccccc|c|cccc}
\hline
\multirow{3}{*}{Method}         & \multirow{3}{*}{Sup.} & \multirow{3}{*}{\# F.} & \multicolumn{6}{c|}{Bucketed Normalized Endpoint Error ($\downarrow$)} & \multicolumn{4}{c}{3-way Endpoint Error ($\downarrow$)} \\ \cline{4-13} 
                                &                       &                        & \multicolumn{5}{c|}{Dynamic} & \multicolumn{1}{c|}{Static} & \multirow{2}{*}{Avg.}   & \multirow{2}{*}{FD}   & \multirow{2}{*}{BS} & \multirow{2}{*}{FS}   \\ \cline{4-9}
                                &                       &                        & mean    & Car             & \makecell{Other \\ vehicles}           & \makecell{Pedes-\\ trian}     & \makecell{Wheeled-\\ vru}         & mean     &              &              \\ \hline
ZeroFlow 1x~\cite{vedder2024zeroflow} &  & 2 & $0.5941$          & $0.3267$          & $0.4756$          & $0.9663$          & $0.6078$          & $0.0195$          & -     & -  & - & -    \\ 
ZeroFlow XL 5x~\cite{vedder2024zeroflow} &  & 2 & $0.4389$          & $0.2382$          & $0.2577$          & $0.7225$          & $0.4517$          & $0.0143$          & $0.0510$     & $0.1239$   & $0.0106$ & $0.0184$   \\ 
NSFP~\cite{li2021NSFP}        &  & 2 & $0.4219$          & $0.2509$          & $0.3313$          & $0.7225$          & $0.3831$          & $0.0279$          & $0.0606$          & $0.1158$       & $0.0344$ & $0.0316$    \\ 
Liu et al.~\cite{liu2024self}        &  & 2 & $0.4134$          & $0.3095$          & $0.5586$          & $0.5092$          & $0.2761$          & $0.0855$          & $0.1307$          & $0.1900$   & $0.1064$ & $0.0956$        \\ 
FastNSF~\cite{li2023fastNSF}        &  & 2 & $0.3826$          & $0.2961$          & $0.4126$          & $0.5002$          & $0.3215$          & $0.0736$          & $0.1118$          & $0.1634$         & $0.0907$ & $0.0814$  \\
ICP-Flow~\cite{lin2024icp}        &  & 2 & $0.3309$          & $0.1945$          & $0.3314$          & $0.4353$          & $0.3626$          & $0.0271$          & $0.0650$          & $0.1369$         & $0.0250$ & $0.0332$  \\
SeFlow~\cite{zhang2024seflow}        &  & 2 & $0.3194$          & $0.2178$          & $0.3464$          & $0.4452$          & $0.2683$          & $0.0148$          & $0.0536$          & $0.1323$         & $0.0043$ & $0.0242$  \\ \hline
FastFlow3D~\cite{jund2022fastflow3d}     & \checkmark & 2 & $0.5323$          & $0.2429$          & $0.3908$          & $0.9818$          & $0.5139$          & $0.0182$          & $0.0735$     & $0.1917$   & $\underline{0.0027}$ & $0.0262$   \\ 
DeFlow~\cite{zhang2024deflow}         & \checkmark & 2 & $0.3704$          & $0.1530$               & $0.3150$               & $0.6615$               & $0.3520$               & $0.0262$               & $0.0501$     & $0.1091$   & $0.0062$ & $0.0352$   \\ 
TrackFlow~\cite{khatri2024can}      & \checkmark & 2 & $0.2689$          & $0.1817$          & $0.3054$          & $0.3581$          & $0.2302$          & $0.0447$          & $0.0473$     & $0.1030$  & $\textbf{0.0024}$ & $0.0365$    \\ \hline
\multirow{2}{*}{Ours} & \checkmark & 2 & $\underline{0.1738}$          & $\underline{0.0954}$          & $\underline{0.1671}$          & $\underline{0.2775}$          & $\underline{0.1554}$          & $\underline{0.0123}$          & $\underline{0.0251}$ & $\underline{0.0573}$ & $0.0030$ & $\underline{0.0148}$ \\  
                                & \checkmark & 5 & $\textbf{0.1454}$ & $\textbf{0.0871}$ & $\textbf{0.1505}$ & $\textbf{0.2165}$ & $\textbf{0.1272}$ & $\textbf{0.0106}$ & $\textbf{0.0224}$ & $\textbf{0.0494}$ & $0.0047$ & $\textbf{0.0130}$ \\ \hline
\end{tabular}}
\label{tab:testset}
\end{table*}

\begin{table*}[t]
\caption{\textbf{Quantitative comparison on the Argoverse 2 validation set.}}
\centering
\begin{tabular}{l|c|cc|cccc|c}
\hline
\multirow{2}{*}{Method} & \multirow{2}{*}{\# Frames} & \multicolumn{2}{c|}{Bucketed Normalized EPE ($\downarrow$)} & \multicolumn{4}{c|}{3-way Endpoint Error ($\downarrow$)} & \multirow{2}{*}{Dynamic IoU ($\uparrow$)} \\ \cline{3-8}
                        &                        & mean Dynamic & mean Static & Avg. & FD & BS & FS & \\ \hline
FastFlow3D~\cite{jund2022fastflow3d} & 2 & $0.6245$ & $0.0146$ & $0.0852$ & $0.2326$ & $\underline{0.0025}$ & $0.0206$ & $0.5257$ \\ \hline
DeFlow~\cite{zhang2024deflow}        & 2 & $0.4305$ & $0.0207$ & $0.0516$ & $0.1212$ & $0.0047$ & $0.0289$ & $0.5227$ \\ \hline
\multirow{2}{*}{Ours}                & 2 & $\underline{0.2047}$ & $\underline{0.0121}$ & $\underline{0.0318}$ & $\underline{0.0773}$ & $\textbf{0.0024}$ & $\underline{0.0157}$ & $\textbf{0.6909}$ \\ 
                                     & 5 & $\textbf{0.1641}$ & $\textbf{0.0104}$ & $\textbf{0.0283}$ & $\textbf{0.0675}$ & $0.0039$ & $\textbf{0.0134}$ & $\underline{0.6278}$ \\  \hline
\end{tabular}
\label{tab:validset}
\end{table*}

\begin{table}[t] 
\caption{\textbf{Comparison of the proposed Spatio-Temporal Decomposition Blocks.}}
\centering
\begin{tabular}{c|cc|c}
\hline
\multicolumn{1}{c|}{\multirow{2}{*}{Module}} & \multicolumn{2}{c|}{Bucketed Normalized EPE ($\downarrow$)} & \multirow{2}{*}{GFLOPs} \\ \cline{2-3}
\multicolumn{1}{c|}{}                         & \multicolumn{1}{c}{Mean Dynamic} & Mean Static & \\ \hline
4D conv                                       & \multicolumn{1}{c}{$\underline{0.1662}$} & $\textbf{0.0097}$ & $152.0$ \\
STDB-B                                        & \multicolumn{1}{c}{$0.1901$} & $0.0115$ & $58.6$ \\
STDB-D                                        & \multicolumn{1}{c}{$0.1719$} & $0.0108$ & $65.1$ \\
STDB-P                                        & \multicolumn{1}{c}{$\textbf{0.1641}$} & $\underline{0.0104}$ & $67.8$ \\ \hline
\end{tabular}
\label{tab:decomposition}
\end{table}

\begin{table}[t]
\caption{\textbf{Performance and computational efficiency based on the number of input frames.}}
\centering
\begin{tabular}{c|cc|cc}
\hline
\multirow{2}{*}{\# Frames} & \multicolumn{2}{c|}{Bucketed Normalized EPE ($\downarrow$)} & \multicolumn{2}{c}{Efficiency} \\ \cline{2-5}
                       & Mean Dynamic & Mean Static & GiB & FPS \\ \hline
2                      & $0.2047$   & $0.0121$  & $1.48$ & $18.9$ \\
3                      & $0.1812$   & $0.0109$  & $1.53$ & $17.8$ \\
4                      & $\underline{0.1712}$   & $\underline{0.0101}$  & $1.57$ & $16.4$ \\
5                      & $\textbf{0.1641}$   & $0.0104$  & $1.62$ & $15.1$ \\
10                      & $0.1728$   & $\textbf{0.0090}$  & $2.00$ & $10.9$ \\ \hline
\end{tabular}
\label{tab:frame}
\end{table}

\begin{table}[t]
\centering
\caption{\textbf{Performance Comparison of Different Component Configurations.} `T. F.' denotes temporal fusion and `B. N. EPE' denotes Bucketed Normalized EPE.}
\label{tab:component}
\resizebox{0.99\columnwidth}{!}{
\begin{tabular}{c|cccc|cc}
\hline
\multirow{2}{*}{Method} & \multicolumn{4}{c|}{Components} & \multicolumn{2}{c}{B. N. EPE ($\downarrow$)} \\ \cline{2-7}
                        & P / V & T. F. & \# Frames & STDB-P & mean D. & mean S. \\ \hline
DeFlow & Pillar & & 2 & & $0.4305$ & $0.0207$ \\ 
(a) & Pillar & \ding{51} & 2 & & $0.2447$ & $0.0127$ \\ 
(b) & Voxel & \ding{51} & 2 & & $0.2084$ & $0.0113$ \\ 
(c) & Voxel & \ding{51} & 5 & & $\underline{0.1662}$ & $\textbf{0.0097}$ \\ 
Flow4D & Voxel & \ding{51} & 5 & \ding{51} & $\textbf{0.1641}$ & $\underline{0.0104}$ \\ \hline
\end{tabular}
}
\end{table}

\begin{table}[]
\caption{\textbf{Computation time for each stage of the Flow4D.}}
\centering
\resizebox{0.99\columnwidth}{!}{
\begin{tabular}{c||c|c|c|c||c}
\hline
Stage & Warping & Voxelization & 4D Network & Head & Total \\ \hline
[ms]   & 0.8     & 18.4          & 46.4       & 0.6  & 66.2 \\ \hline
\end{tabular}
}
\label{tab:computation}
\end{table}

% Experiments

\section{Experiments}
In this section, we demonstrate the superiority of the proposed Flow4D in the scene flow task.
In~\Secref{sec:Experiment_detail}, we describe the dataset, metrics, and training details used in the experiments.
Then, we quantitatively and qualitatively compare the proposed method with other state-of-the-art methods in~\Secref{sec:comparison}.
% Finally, in~\Secref{sec:ablation}, we validate the effectiveness of each module of Flow4D through ablation studies, including detailed analysis of the computational load.
In~\Secref{sec:ablation}, we validate the effectiveness of each module of Flow4D through ablation studies. Finally, analysis of the computational time is discussed in~\Secref{sec:computation}.
% This section outlines the comprehensive experimentation conducted to validate the effectiveness of our method. 
% We detail the model implementation and experimental setups in~\Secref{sec:detail}. 
% We compare our method against existing condition-free domain generalization techniques in~\Secref{sec:condition-free}. 
% % We evaluate it alongside condition-required domain adaptation and generalization methods in~\secref{sec:condition-required}.
% We evaluate the proposed method alongside condition-required domain adaptation and generalization methods in~\Secref{sec:condition-required}. 
% Subsequently, \Secref{sec:ablation} is dedicated to dissecting the influence of each component within our framework.
% Throughout these experiments, we adhere to the class map settings established in prior studies to ensure our evaluations are consistent with recognized benchmarks.
% Additionally, \Secref{sec:speed} investigates the computational demands of our approach.

% Table 사용?

\subsection{Experimental Settings}
\label{sec:Experiment_detail}
\noindent\textbf{Dataset:} We utilize the large-scale Argoverse 2 dataset~\cite{Benjamin2021Argoverse2} to evaluate the proposed method. The Argoverse 2 dataset comprises 110,071 training data and 23,573 test data.
Ground truth is not provided for the test set, and the evaluation for the test set is conducted on the official benchmark server.
Since Argoverse 2 is the only LiDAR-based scene flow dataset with a blind test server, we focus experiments on the Argoverse 2.

\noindent\textbf{Metrics:}
The Argoverse 2 2024 scene flow challenge uses Bucket Normalized Endpoint Error (EPE)~\cite{khatri2024can} as the main metric, while the Argoverse 2 2023 scene flow challenge used Three-way EPE~\cite{chodosh2024Re-eval} as the main metric.
Following this, we use the Bucket Normalized EPE and the Three-way EPE as main metrics.
Due to the severe class imbalance in LiDAR point clouds, simply using average EPE tends to focus on the EPE of static objects.
To address this, Three-way EPE calculates the EPE for dynamic foreground (FD), static foreground (FS), and static background (BS) separately and then averages them.
Furthermore, Bucket Normalized EPE divides classes into background, car, other-vehicle, pedestrian, and wheeled-VRU, and normalizes errors based on the speed of the objects, enabling a more detailed and accurate performance comparison.
% However, Three-way EPE still fails to account for the performance on small objects and treats errors equally regardless of the object's speed.
%Bucketed Normalized EPE, Three-way EPE 설명 적는다?

% \subsection{Training Details}
% \label{sec:details}

\noindent\textbf{Training Details:}
We train the model using the DeFlow loss~\cite{zhang2024deflow}, which effectively scales dynamic and static points, with the Adam optimizer (lr=1e-3) for 15 epochs.
% We use five consecutive frames from \(t-3\) to \(t+1\) as input.
In the voxelization process, each voxel size is \([20 \text{cm}, 20 \text{cm}, 20 \text{cm}]\), and the voxelization range is \([51.2\text{m}, 51.2\text{m}, 6.4\text{m}]\).
Finally, the resolution of the 4D voxel space is \([W=512, L=512, H=32, T=5]\).

\subsection{Comparison}
\label{sec:comparison}

\begin{figure*}[t] 
    \centering
    \includegraphics[width=0.99\linewidth]{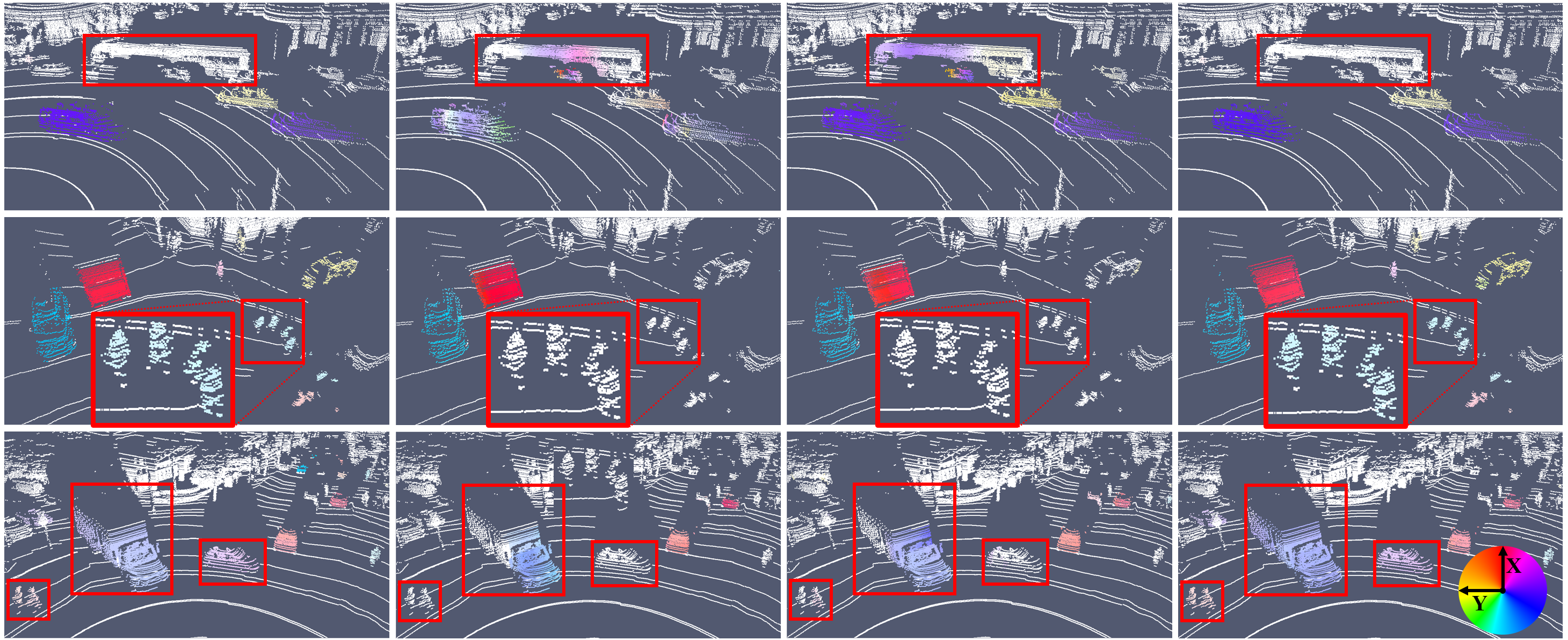}

    %\vspace{2mm} % 이미지와 텍스트 사이의 간격 조절
    \begin{tabularx}{\textwidth}{XXXX}
        \centering Ground-truth & \centering FastFlow3D~\cite{jund2022fastflow3d} & \centering DeFlow~\cite{zhang2024deflow} & \centering Ours
    \end{tabularx}

    \caption{Qualitative comparison on Argoverse 2 validation set.
    Following DeFlow~\cite{zhang2024deflow}, we represent the direction and magnitude of each motion vector as hue and saturation, respectively.
    Our proposed method estimates motion vectors more accurately than existing methods~\cite{jund2022fastflow3d, zhang2024deflow} for all classes, including pedestrians, vehicles, trucks, and buses.}
    \label{figure_qualitative}
    
    % \vspace{2mm} % 이미지와 텍스트 사이의 간격 조절
    % \parbox{\linewidth}{
    %     \centering
    %     Ground-truth \hspace{2.5cm} FastFlow3D~\cite{jund2022fastflow3d} \hspace{2.5cm} DeFlow~\cite{zhang2024deflow} \hspace{3cm} Ours
    % }
\end{figure*}

We compare the proposed method with various supervised and self-supervised LiDAR scene flow methods. \tabref{tab:testset} shows the comparison results evaluated through the official test server of Argoverse 2.
The proposed method achieves a mean Dynamic Normalized EPE of 0.1454, which is 45.9\% better than the state-of-the-art method, TrackFlow~\cite{khatri2024can}, and 60.7\% better than DeFlow~\cite{zhang2024deflow}, demonstrating remarkable performance.
Additionally, the proposed method outperforms all existing methods in class-specific Dynamic Normalized EPE and mean Static Normalized EPE.
%Test server에 3-way EPE 열리면 내용 추가하기.
Even when using only 2 frames, the proposed method surpasses TrackFlow by 35.4\% in mean Dynamic Normalized EPE.
Additionally, as shown in \tabref{tab:validset}, we further compare the proposed method with supervised methods FastFlow3D~\cite{jund2022fastflow3d} and DeFlow~\cite{zhang2024deflow} on the Argoverse 2 validation set.
We define dynamic points using a 0.5 m/s threshold and compare the Dynamic IoU.
When using 2 frames, our method achieves a 52.5\% lower mean Dynamic Normalized EPE compared to DeFlow.
Furthermore, using 5 frames results in an additional 19.8\% performance improvement over using 2 frames.
These results demonstrate the superiority of the proposed Flow4D.
Additionally, we present qualitative comparisons in \figref{figure_qualitative}. As shown in \figref{figure_qualitative}, Flow4D performs more accurate motion vectors for all classes compared to FastFlow3D~\cite{jund2022fastflow3d} and DeFlow~\cite{zhang2024deflow}.
% For more detailed results of our method, please refer to the attached video in the multimedia attachment.
For more detailed results, please refer to the multimedia attachment.
% For more detailed results, please refer to the attached video.
%{figure_qualitative}

\subsection{Ablation Studies}
\label{sec:ablation}
To provide a comprehensive analysis of the contributions of each element, we conduct further experiments to evaluate the proposed Spatio-Temporal Decomposition Blocks (STDB), the impact of varying the number of input frames, and the effectiveness of each component on the Argoverse 2 validation set.

\noindent\textbf{Spatio-Temporal Decomposition Block:}
% We compare the performance and efficiency of the proposed Decomposition blocks and analyze the impact of the number of frames used.
% \tabref{tab:decomposition} shows the performance comparison between the 4D convolution residual block and the proposed Spatio-temporal Decomposition Block (STDB).
\tabref{tab:decomposition} shows the performance comparison between the 4D convolution residual block (\figref{fig_decomposition}a), the Baseline STDB (\figref{fig_decomposition}b), the Dual-path STDB (\figref{fig_decomposition}c), and the Parallel STDB (\figref{fig_overview}).
The STDB-Baseline requires 61.4\% less GFLOPs compared to using the 4D convolution, but suffers a 14.4\% increase in mean Dynamic EPE.
The STDB-Dual-path consumes 11.1\% more GFLOPs compared to the STDB-B but achieves a 9.6\% reduction in mean Dynamic EPE.
Lastly, the STDB-Parallel has a 16.6\% higher computational load compared to the STDB-B, but shows a 55.4\% lower computational load compared to using the 4D Conv while achieving superior mean Dynamic EPE.
% In contrast, the Parallel spatio-temporal block demonstrates a 27.0\% reduction in GPU memory usage while achieving a 1.3\% improvement in performance.
% Although the Dual-path block showed the best computation speed, we selected the Parallel block as the base module for the proposed method considering the balance between performance and efficiency.

% \tabref{tab:decomposition} shows the performance comparison between the 4D convolution-based residual block and the proposed Spatio-temporal Decomposition Blocks (STDB).
% The simple Sequential spatio-temporal block uses 8.1\% less memory and has a 40.6\% faster computation speed compared to the 4D conv, but suffers a 4.8\% performance degradation.
% In contrast, the Parallel spatio-temporal block demonstrates a 27.0\% reduction in GPU memory usage and a 9.8\% increase in computation speed while achieving a 1.3\% improvement in performance.
% Although the Dual-path block showed the best computation speed, we selected the Parallel block as the base module for the proposed method considering the balance between performance and efficiency.

%Dual-path
%Parallel

\noindent\textbf{Number of frames:}
We analyze the performance and efficiency based on the number of frames used, as shown in \tabref{tab:frame}.
Increasing the number of frames tends to improve performance, but also increases the computational overhead.
The proposed method outperforms the current state-of-the-art even with just two frames, making it advisable to use two frames in scenarios where efficiency is a priority.
When using five frames, Flow4D achieves the lowest mean Dynamic EPE, and when using ten frames, it achieves the lowest mean Static EPE.
Thanks to the early-fusion strategy, the proposed method does not experience a significant increase in computational overhead with more frames, and it can operate in real-time using only 1.62GiB of GPU memory when using five frames.
Considering both performance and efficiency, we employ five frames as the default setting for the proposed Flow4D.

\noindent\textbf{Effectiveness of each component:} 
\tabref{tab:component} shows the effectiveness of each component of the proposed method.
DeFlow~\cite{zhang2024deflow} converts two point clouds into 2D BEV and applies individual 2D encoders.
% Method (a) retains the pillar-based representation but stacks the 2D BEV features \([W, L, 16]\) along the time axis to create a 4D tensor \([W, L, 2, 16]\) and utilizes the 3D voxel network, reducing the mean Dynamic to 0.2447 ($\downarrow 43.2\%$).
Method (a) retains the pillar-based representation but stacks the 2D BEV features \([W, L, 16]\), obtained from the pillar feature encoder, along the time axis to create a 4D tensor \([W, L, 2, 16]\).
Then, it utilizes the 3D voxel network, reducing the mean Dynamic EPE to 0.2447 ($\downarrow 43.2\%$).
% Method (a) retains the pillar-based representation but stacks the 2D BEVs \([W, L, f]\) along the time axis to create a 4D tensor \([W, L, T, f]\) and utilizes the 3D voxel-based network, reducing the mean Dynamic to 0.2447 ($\downarrow 43.2\%$).
Method (b) transitions from pillar-based to a voxel-based representation, resulting in a mean Dynamic EPE of 0.2084 ($\downarrow 14.8\%$ compared to (a)).
Increasing the number of input frames from 2 to 5 (method (c)) results in an additional 20.2\% performance improvement.
Finally, the application of the Spatio-Temporal Decomposition Block (STDB) results in an additional 1.3\% performance improvement.
Incorporating all these components, the proposed Flow4D shows a 61.9\% performance improvement over DeFlow~\cite{zhang2024deflow}.

% Stacking the 2D BEVs \([W, L, f]\) in the time channel to create a 4D tensor \([W, L, T, f]\) and using 3D Sparse U-Net results in a 43.2\% performance improvement over DeFlow.

\subsection{Computational Time}
\label{sec:computation}
\tabref{tab:computation} shows the computation time required for each component of Flow4D.
The proposed method takes a total of 66.2 ms from point cloud warping to output on the NVIDIA RTX 3090.
Among the total computation time, the 4D voxel network takes 46.4 ms, and the voxelization process takes 18.4 ms.
The proposed Flow4D processes each of the five input frames through 3D voxelization, contributing some computational overhead during the voxelization stage.
Nevertheless, the proposed method operates at 15.1 FPS, making real-time operation feasible in autonomous driving systems.

\section{Conclusion}
In this paper, we propose Flow4D, a novel 4D voxel-based LiDAR scene flow framework.
The proposed method takes five 3D point clouds and converts them into 3D voxels using an intra-voxel feature encoder (VFE).
Then, it expands the dimension of each 3D voxel by adding a time dimension and concatenates them along this axis to create a 4D voxel.
Through the 4D voxel network, it explicitly extracts spatial and temporal information simultaneously.
To reduce the heavy computational load of 4D convolution, we introduce the Spatio-Temporal Decomposition Block (STDB), which decomposes 4D convolution into 3D spatial convolution and 1D temporal convolution.
Furthermore, Flow4D is designed with an early fusion scheme that fuses features immediately after the VFE, allowing it to effectively exploit more temporal cues from a larger number of point clouds.
These contributions enable our method to achieve a significant performance improvement of 45.9\% over the previous state-of-the-art while being efficient enough to run in real-time on a desktop GPU.
In the future, we plan to develop an even more efficient framework capable of real-time operation on embedded systems.

{\small
\bibliographystyle{IEEEtran}
\bibliography{egbib}

\begin{thebibliography}{10}
\providecommand{\url}[1]{#1}
\csname url@rmstyle\endcsname
\providecommand{\newblock}{\relax}
\providecommand{\bibinfo}[2]{#2}
\providecommand\BIBentrySTDinterwordspacing{\spaceskip=0pt\relax}
\providecommand\BIBentryALTinterwordstretchfactor{4}
\providecommand\BIBentryALTinterwordspacing{\spaceskip=\fontdimen2\font plus
\BIBentryALTinterwordstretchfactor\fontdimen3\font minus \fontdimen4\font\relax}
\providecommand\BIBforeignlanguage[2]{{%
\expandafter\ifx\csname l@#1\endcsname\relax
\typeout{** WARNING: IEEEtran.bst: No hyphenation pattern has been}%
\typeout{** loaded for the language `#1'. Using the pattern for}%
\typeout{** the default language instead.}%
\else
\language=\csname l@#1\endcsname
\fi
#2}}

\bibitem{li2023pillarnext}
J.~Li, C.~Luo, and X.~Yang, ``Pillarnext: Rethinking network designs for 3d object detection in lidar point clouds,'' in \emph{Proceedings of IEEE Conference on Computer Vision and Pattern Recognition (CVPR)}, 2023, pp. 17\,567--17\,576.

\bibitem{liu2024multi}
Y.~Liu, L.~Kong, X.~Wu, R.~Chen, X.~Li, L.~Pan, Z.~Liu, and Y.~Ma, ``Multi-space alignments towards universal lidar segmentation,'' in \emph{Proceedings of IEEE Conference on Computer Vision and Pattern Recognition (CVPR)}, 2024, pp. 14\,648--14\,661.

\bibitem{jund2022fastflow3d}
P.~Jund, C.~Sweeney, N.~Abdo, Z.~Chen, and J.~Shlens, ``Scalable scene flow from point clouds in the real world,'' \emph{IEEE Robotics and Automation Letters}, vol.~7, no.~2, pp. 1589--1596, 2022.

\bibitem{zhang2024deflow}
Q.~Zhang, Y.~Yang, H.~Fang, R.~Geng, and P.~Jensfelt, ``Deflow: Decoder of scene flow network in autonomous driving,'' in \emph{Proceedings of IEEE International Conference on Robotics and Automation (ICRA)}.\hskip 1em plus 0.5em minus 0.4em\relax IEEE, 2024.

\bibitem{li2021NSFP}
X.~Li, J.~Kaesemodel~Pontes, and S.~Lucey, ``Neural scene flow prior,'' \emph{Advances in Neural Information Processing Systems}, vol.~34, pp. 7838--7851, 2021.

\bibitem{li2023fastNSF}
X.~Li, J.~Zheng, F.~Ferroni, J.~K. Pontes, and S.~Lucey, ``Fast neural scene flow,'' in \emph{Proceedings of IEEE International Conference on Computer Vision (ICCV)}, 2023, pp. 9878--9890.

\bibitem{liu2024self}
D.~Liu, D.~Liu, X.~Li, S.~Lin, B.~Wang, X.~Chang, L.~Chu, \emph{et~al.}, ``Self-supervised multi-frame neural scene flow,'' \emph{arXiv preprint arXiv:2403.16116}, 2024.

\bibitem{vedder2024zeroflow}
K.~Vedder, N.~Peri, N.~E. Chodosh, I.~Khatri, E.~Eaton, D.~Jayaraman, Y.~Liu, D.~Ramanan, and J.~Hays, ``Zeroflow: Scalable scene flow via distillation,'' in \emph{International Conference on Learning Representations (ICLR)}, 2024.

\bibitem{lin2024icp}
Y.~Lin and H.~Caesar, ``Icp-flow: Lidar scene flow estimation with icp,'' in \emph{Proceedings of IEEE Conference on Computer Vision and Pattern Recognition (CVPR)}, 2024.

\bibitem{khatri2024can}
I.~Khatri, K.~Vedder, N.~Peri, D.~Ramanan, and J.~Hays, ``I can't believe it's not scene flow!'' \emph{arXiv preprint arXiv:2403.04739}, 2024.

\bibitem{Benjamin2021Argoverse2}
B.~Wilson, W.~Qi, T.~Agarwal, J.~Lambert, J.~Singh, S.~Khandelwal, B.~Pan, R.~Kumar, A.~Hartnett, J.~K. Pontes, D.~Ramanan, P.~Carr, and J.~Hays, ``Argoverse 2: Next generation datasets for self-driving perception and forecasting,'' in \emph{Proceedings of the Neural Information Processing Systems Track on Datasets and Benchmarks (NeurIPS Datasets and Benchmarks 2021)}, 2021.

\bibitem{Lang2019Pointpillars}
A.~H. Lang, S.~Vora, H.~Caesar, L.~Zhou, J.~Yang, and O.~Beijbom, ``Pointpillars: Fast encoders for object detection from point clouds,'' in \emph{Proceedings of IEEE Conference on Computer Vision and Pattern Recognition (CVPR)}, 2019, pp. 12\,697--12\,705.

\bibitem{Cortinhal2020SalsaNext}
T.~Cortinhal, G.~Tzelepis, and E.~Erdal~Aksoy, ``Salsanext: Fast, uncertainty-aware semantic segmentation of lidar point clouds,'' in \emph{Advances in Visual Computing: 15th International Symposium, ISVC 2020, San Diego, CA, USA, October 5--7, 2020, Proceedings, Part II 15}.\hskip 1em plus 0.5em minus 0.4em\relax Springer, 2020, pp. 207--222.

\bibitem{Ando2023Rangevit}
A.~Ando, S.~Gidaris, A.~Bursuc, G.~Puy, A.~Boulch, and R.~Marlet, ``Rangevit: Towards vision transformers for 3d semantic segmentation in autonomous driving,'' in \emph{Proceedings of IEEE Conference on Computer Vision and Pattern Recognition (CVPR)}, 2023, pp. 5240--5250.

\bibitem{Thomas2019KPConv}
H.~Thomas, C.~R. Qi, J.-E. Deschaud, B.~Marcotegui, F.~Goulette, and L.~J. Guibas, ``Kpconv: Flexible and deformable convolution for point clouds,'' in \emph{Proceedings of IEEE International Conference on Computer Vision (ICCV)}, 2019, pp. 6411--6420.

\bibitem{Hu2020RandLA}
Q.~Hu, B.~Yang, L.~Xie, S.~Rosa, Y.~Guo, Z.~Wang, N.~Trigoni, and A.~Markham, ``Randla-net: Efficient semantic segmentation of large-scale point clouds,'' in \emph{Proceedings of IEEE Conference on Computer Vision and Pattern Recognition (CVPR)}, 2020, pp. 11\,108--11\,117.

\bibitem{Choyi2019MINKOWSKI}
C.~Choy, J.~Gwak, and S.~Savarese, ``4d spatio-temporal convnets: Minkowski convolutional neural networks,'' in \emph{Proceedings of IEEE Conference on Computer Vision and Pattern Recognition (CVPR)}, 2019, pp. 3075--3084.

\bibitem{Zhu2021Cylindrical}
X.~Zhu, H.~Zhou, T.~Wang, F.~Hong, Y.~Ma, W.~Li, H.~Li, and D.~Lin, ``Cylindrical and asymmetrical 3d convolution networks for lidar segmentation,'' in \emph{Proceedings of IEEE Conference on Computer Vision and Pattern Recognition (CVPR)}, 2021, pp. 9939--9948.

\bibitem{liu2015sparse}
B.~Liu, M.~Wang, H.~Foroosh, M.~Tappen, and M.~Pensky, ``Sparse convolutional neural networks,'' in \emph{Proceedings of IEEE Conference on Computer Vision and Pattern Recognition (CVPR)}, 2015, pp. 806--814.

\bibitem{graham2017submanifold}
B.~Graham and L.~Van~der Maaten, ``Submanifold sparse convolutional networks,'' \emph{arXiv preprint arXiv:1706.01307}, 2017.

\bibitem{mittal2020just}
H.~Mittal, B.~Okorn, and D.~Held, ``Just go with the flow: Self-supervised scene flow estimation,'' in \emph{Proceedings of the IEEE/CVF conference on computer vision and pattern recognition}, 2020, pp. 11\,177--11\,185.

\bibitem{pontes2020scene}
J.~K. Pontes, J.~Hays, and S.~Lucey, ``Scene flow from point clouds with or without learning,'' in \emph{2020 international conference on 3D vision (3DV)}.\hskip 1em plus 0.5em minus 0.4em\relax IEEE, 2020, pp. 261--270.

\bibitem{kittenplon2021flowstep3d}
Y.~Kittenplon, Y.~C. Eldar, and D.~Raviv, ``Flowstep3d: Model unrolling for self-supervised scene flow estimation,'' in \emph{Proceedings of the IEEE/CVF Conference on Computer Vision and Pattern Recognition}, 2021, pp. 4114--4123.

\bibitem{li2022rigidflow}
R.~Li, C.~Zhang, G.~Lin, Z.~Wang, and C.~Shen, ``Rigidflow: Self-supervised scene flow learning on point clouds by local rigidity prior,'' in \emph{Proceedings of IEEE Conference on Computer Vision and Pattern Recognition (CVPR)}, 2022, pp. 16\,959--16\,968.

\bibitem{zhang2024seflow}
Q.~Zhang, Y.~Yang, P.~Li, O.~Andersson, and P.~Jensfelt, ``Seflow: A self-supervised scene flow method in autonomous driving,'' \emph{arXiv preprint arXiv:2407.01702}, 2024.

\bibitem{behl2019pointflownet}
A.~Behl, D.~Paschalidou, S.~Donn{\'e}, and A.~Geiger, ``Pointflownet: Learning representations for rigid motion estimation from point clouds,'' in \emph{Proceedings of the IEEE/CVF Conference on Computer Vision and Pattern Recognition}, 2019, pp. 7962--7971.

\bibitem{liu2019flownet3d}
X.~Liu, C.~R. Qi, and L.~J. Guibas, ``Flownet3d: Learning scene flow in 3d point clouds,'' in \emph{Proceedings of the IEEE/CVF conference on computer vision and pattern recognition}, 2019, pp. 529--537.

\bibitem{puy2020flot}
G.~Puy, A.~Boulch, and R.~Marlet, ``Flot: Scene flow on point clouds guided by optimal transport,'' in \emph{European conference on computer vision}.\hskip 1em plus 0.5em minus 0.4em\relax Springer, 2020, pp. 527--544.

\bibitem{wu2020pointpwc}
W.~Wu, Z.~Y. Wang, Z.~Li, W.~Liu, and L.~Fuxin, ``Pointpwc-net: Cost volume on point clouds for (self-) supervised scene flow estimation,'' in \emph{Computer Vision--ECCV 2020: 16th European Conference, Glasgow, UK, August 23--28, 2020, Proceedings, Part V 16}.\hskip 1em plus 0.5em minus 0.4em\relax Springer, 2020, pp. 88--107.

\bibitem{lee2020pillarflow}
K.-H. Lee, M.~Kliemann, A.~Gaidon, J.~Li, C.~Fang, S.~Pillai, and W.~Burgard, ``Pillarflow: End-to-end birds-eye-view flow estimation for autonomous driving,'' in \emph{2020 IEEE/RSJ International Conference on Intelligent Robots and Systems (IROS)}.\hskip 1em plus 0.5em minus 0.4em\relax IEEE, 2020, pp. 2007--2013.

\bibitem{li2021hcrf}
R.~Li, G.~Lin, T.~He, F.~Liu, and C.~Shen, ``Hcrf-flow: Scene flow from point clouds with continuous high-order crfs and position-aware flow embedding,'' in \emph{Proceedings of the IEEE/CVF Conference on Computer Vision and Pattern Recognition}, 2021, pp. 364--373.

\bibitem{battrawy2022rms}
R.~Battrawy, R.~Schuster, M.-A.~N. Mahani, and D.~Stricker, ``Rms-flownet: Efficient and robust multi-scale scene flow estimation for large-scale point clouds,'' in \emph{Proceedings of IEEE International Conference on Robotics and Automation (ICRA)}.\hskip 1em plus 0.5em minus 0.4em\relax IEEE, 2022.

\bibitem{chodosh2024Re-eval}
N.~Chodosh, D.~Ramanan, and S.~Lucey, ``Re-evaluating lidar scene flow for autonomous driving,'' in \emph{Proceedings of IEEE Winter Conference on Applications of Computer Vision (WACV)}, 2023.

\end{thebibliography}
}

\end{document}